\documentclass{article} 
\usepackage{iclr2025_conference,times}

\usepackage{amsmath,amsfonts,bm}

\def\eqref#1{equation~\ref{#1}}

\def\1{\bm{1}}

\DeclareMathAlphabet{\mathsfit}{\encodingdefault}{\sfdefault}{m}{sl}
\SetMathAlphabet{\mathsfit}{bold}{\encodingdefault}{\sfdefault}{bx}{n}

\usepackage{graphicx}
\usepackage{bbding}
\usepackage[normalem]{ulem}
\usepackage{multirow}
\usepackage{xcolor}
\usepackage{mdframed,lipsum}
\usepackage{hyperref}
\usepackage{url}

\title{Towards democratizing Multilingual Large Language Models for Medicine through a Two-Stage Instruction Fine-tuning Approach\thanks{Work in Progress}}

\author{Meng Zhou, Surajsinh Parmar, Anubhav Bhatti \\
AI Engineering Team\\
SpassMed Inc.\\
Toronto, Ontario, Canada\\
\texttt{\{simon.zhou,suraj.parmar,anubhav.bhatti\}@spassmed.ca}
}

\newcommand*{\escape}[1]{\texttt{\textbackslash#1}}

\iclrfinalcopy 
\begin{document}

\maketitle

\begin{abstract}
Open-source, multilingual medical large language models (LLMs) have the potential to serve linguistically diverse populations across different regions. Adapting generic LLMs for healthcare often requires continual pretraining, but this approach is computationally expensive and sometimes impractical. Instruction fine-tuning on a specific task may not always guarantee optimal performance due to the lack of broader domain knowledge that the model needs to understand and reason effectively in diverse scenarios. To address these challenges, we introduce two multilingual instruction fine-tuning datasets, MMed-IFT and MMed-IFT-MC, containing over 200k high-quality medical samples in six languages. We propose a two-stage training paradigm: the first stage injects general medical knowledge using MMed-IFT, while the second stage fine-tunes on task-specific multiple-choice questions with MMed-IFT-MC. Our method achieves competitive results on both English and multilingual benchmarks, striking a balance between computational efficiency and performance. We plan to make our dataset and model weights public at \url{https://github.com/SpassMed/Med-Llama3} in the future.
\end{abstract}

\section{Introduction}

Large language models (LLMs), such as GPT-4 \citep{achiam2023gpt} and Gemini \citep{reid2024gemini}, have shown remarkable capabilities across various applications. However, their deployment in healthcare is limited by the need for extensive domain-specific knowledge and raises significant concerns related to their proprietary nature, as well as issues around privacy, stability, and transparency. Hospitals and other medical institutions often require models capable of understanding and explaining complex clinical terminology, reasoning through medical scenarios, and providing accurate, context-appropriate responses in different languages. These requirements are critical for ensuring that LLMs can support worldwide medical professionals effectively and safely in real-world clinical environments. Another issue is the significant performance gap between proprietary (closed-source) and open-source models. For example, Med-PaLM 2 \citep{singhal2023towards}, a closed-source medical LLM developed by Google, demonstrates excellent performance across various medical benchmarks. However, its closed nature prevents institutions from conducting rigorous evaluation and validation on private datasets, limiting its adaptability in sensitive healthcare settings. Additionally, sending healthcare data to private companies for model optimization poses privacy risks and operational challenges. This underscores the importance of institutions having the ability to serve and optimize their own models, ensuring reliable and safe access to LLMs tailored to their specific needs. 

Recently, there have been efforts to develop open-source models for biomedicine and healthcare. MedAlpaca \citep{han2023medalpaca} introduced a series of Llama-based \citep{touvron2023llama} open-source models fine-tuned on 16,000 curated English medical conversational datasets. Similarly, ChatDoctor \citep{li2023chatdoctor} is another series of Llama-based models trained on over 100,000 real patient-doctor dialogues in English. MedInstruct \citep{zhang2023alpacare} developed a 54k hand-crated medical instruction-response dataset for instruction fine-tuning Llama-based models. ClinicalCamel \citep{toma2023clinical} used 100k synthetic medical conversations, which were generated by translating 20k clinical articles, along with MedQA \citep{jin2021disease} and data from ShareGPT\footnote{\url{https://sharegpt.com/}} to fine-tuned Llama2-13B and 70B models. Another line of works, such as those by Meditron \citep{chen2023meditron}, MMed \citep{qiu2024towards}, and PMC-Llama \citep{wu2024pmc}, have focused on \textit{continually pre-training} base models to enhance their understanding of fundamental medical knowledge, followed by instruction fine-tuning on specific medical tasks. However, this approach has a significant drawback: continual pre-training demands vast computational resources, as it involves optimizing the entire model with the same pre-trained task using the medical corpus. This makes it impractical for institutions with limited computational power. 

Multilingual medical large language models (LLMs) have garnered increasing attention recently, as the demand for medical knowledge across diverse languages continues to grow. Despite advancements in open-source medical LLMs, most remain monolingual, primarily focusing on English, which limits their accessibility to a linguistically diverse global audience. Recent efforts have begun to address this limitation with bilingual and multilingual medical LLMs. For example, Tianyi \citep{luo2024taiyi} is a bilingual model trained on 10 English and Chinese biomedical tasks using over 1 million curated samples, while BiMediX \citep{pieri2024bimedix} is an English-Arabic LLM based on Mistral 7B \citep{jiang2023mistral}, fine-tuned with 1.3 million multiple-choice and open-ended question-answer samples. Additionally, Apollo \citep{wang2024apollo} and MMed \citep{qiu2024towards} are advanced multilingual medical LLMs covering six languages and trained on billions of tokens.

This paper explores a computationally efficient approach for fine-tuning LLMs while maintaining high performance. To achieve this, we construct two \textit{multilingual instruction fine-tuning datasets}, MMed-IFT and MMed-IFT-MC, designed for general medical knowledge injection and task-specific fine-tuning, respectively. The fine-tuning process for the Llama3 base model is split into two stages, utilizing parameter-efficient fine-tuning (PEFT) with LoRA \citep{hu2021lora} in both phases.

\begin{itemize}
    \item We introduce MMed-IFT and MMed-IFT-MC datasets, containing over 200k high-quality samples from English, Chinese, Japanese, Korean, French, and Spanish.
    \item We introduce a series of preprocessing techniques, including question-answer pair generation and cycle-consistency translation criteria to ensure high-quality data obtained in our datasets.
    \item We propose a two-stage fine-tuning paradigm: general medical knowledge is injected during the first stage using the MMed-IFT dataset, and task-specific fine-tuning is performed in the second stage using the MMed-IFT-MC dataset. Both stages leverage LoRA for parameter-efficient fine-tuning (PEFT), offering a computationally efficient alternative to continual pretraining while maintaining good performance across multilingual medical benchmarks.
\end{itemize}

\section{Methodology}

\begin{figure}[ht] 
	\begin{center}
		\includegraphics[width=\textwidth]{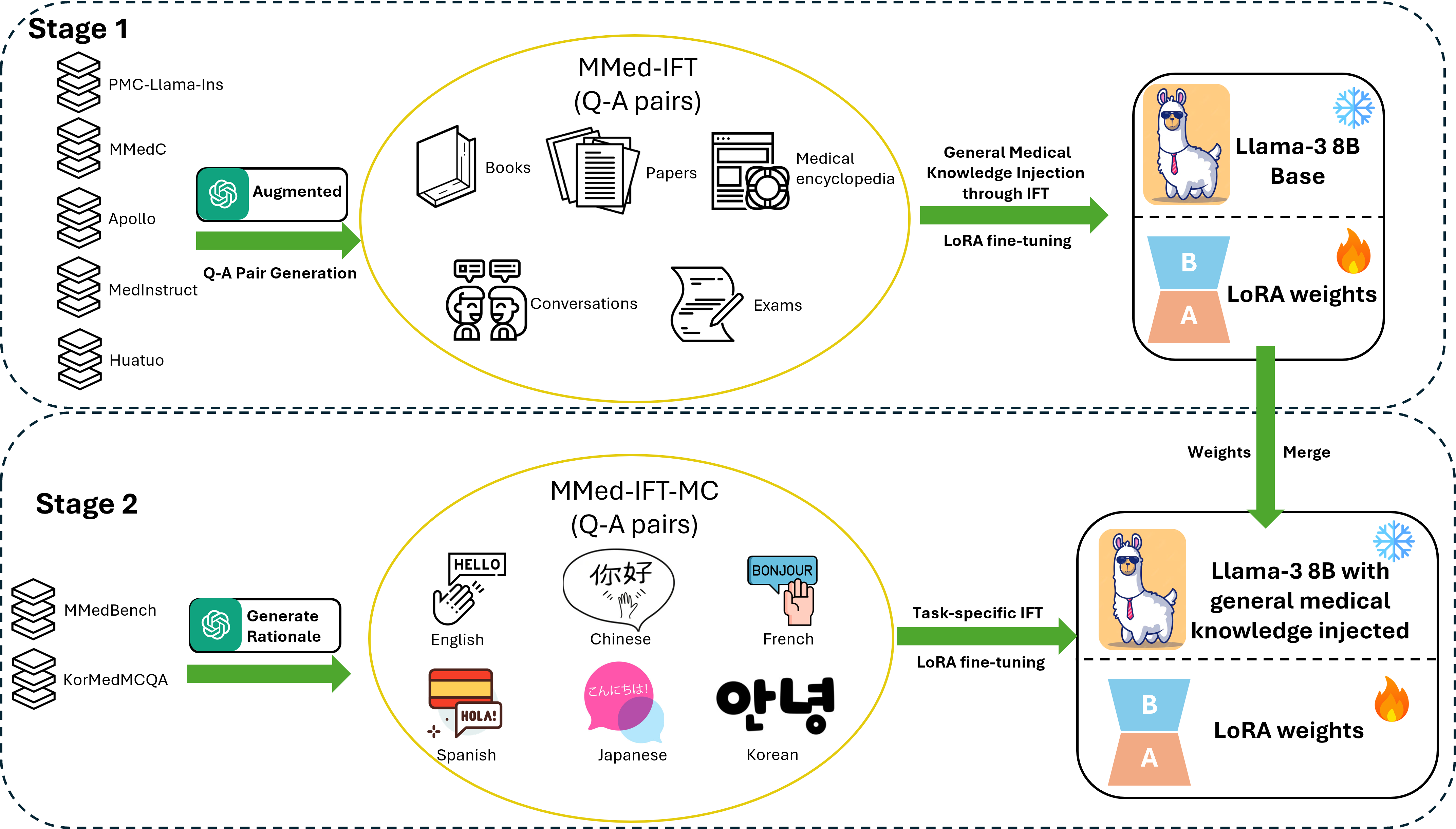}
	\end{center}
	\caption{An overview of this work. Our MMed-IFT dataset is based on various GPT-augmented open-resourced datasets and our hand-crafted samples. The MMed-IFT-MC dataset is based on public MMedBench and KorMedMCQA datasets with GPT-generated rationale. MMed-IFT dataset contains a wide range of medical questions presented in the question-answer format, which is used in Stage 1. MMed-IFT-MC is a more task-specific dataset tailored for Medical Licenses Examination-style multiple-choice questions, which is used in Stage 2.}
	\label{overall}
\end{figure}

In this section, we provide details of our methodology as shown in Figure \ref{overall}. Specifically, we first introduce the construction pipeline for our MMed-IFT and MMed-IFT-MC datasets. Then, we provide details of our proposed two-stage instruction fine-tuning strategy. Finally, we talk about the baseline models used and benchmark evaluations in this work.

\subsection{MMed-IFT and MMed-IFT-MC Dataset}

\begin{figure}[ht] 
	\begin{center}
		\includegraphics[width=\textwidth]{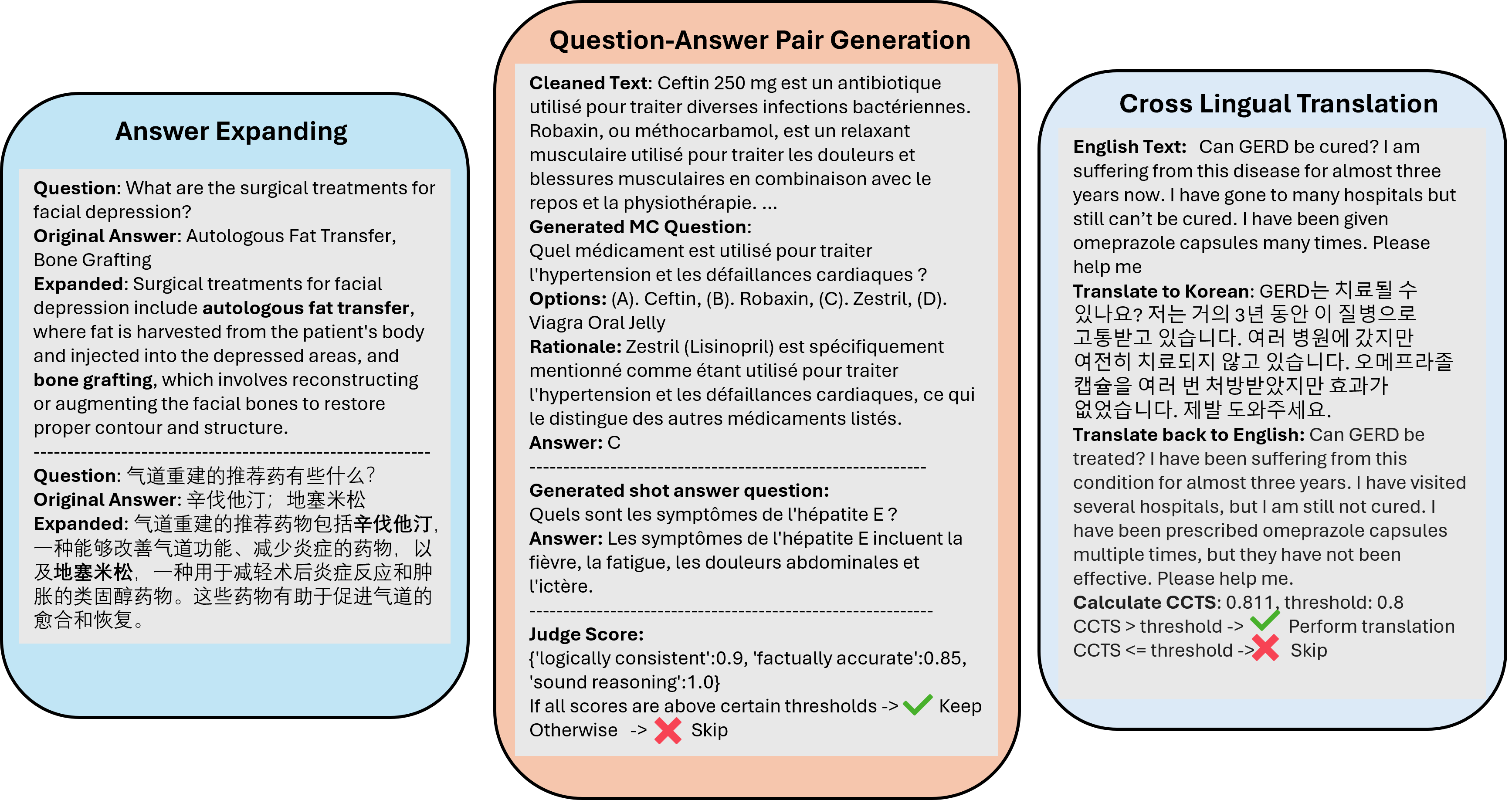}
	\end{center}
	\caption{An example of preprocessing methods we utilized in this work. From left to right: answer expanding, question-answer pair generation, and cross-lingual translation.}
	\label{preprocess}
\end{figure}

In this work, we present \textbf{MMed-IFT} and \textbf{MMed-IFT-MC}, two new large-scale multilingual medical instruction fine-tuning (IFT) datasets designed to enhance large language models (LLMs) with both general and task-specific medical knowledge across multiple languages, while maintaining computational efficiency. Unlike MMedC \citep{qiu2024towards}, which is used for auto-regressive continual pretraining of LLMs, our dataset is specifically developed for parameter-efficient instruction fine-tuning, making it ideal for use in environments with limited computational resources. The MMed-IFT dataset incorporates a variety of publicly available and hand-crafted datasets, covering \textbf{six main languages:} English, Chinese, Japanese, Korean, French, and Spanish. For each language, we adapt its pre-training corpus into a question-answer pair format, integrating it with existing question-answer datasets and our hand-crafted datasets. In the following sections, we will first outline the preprocessing methods applied to clean and generate our datasets, and provide graphical illustrations of some preprocessing methods in Figure \ref{preprocess}, followed by a detailed description of the datasets used for each language.

\noindent \textbf{Knowledge Density Filtering. }This filtering method ensures that samples are relevant to the medical domain, similar to the method used in \cite{qiu2024towards}. Following their methodology, we utilize a carefully curated list of 200 medical terms as reference keywords. For each sample, we perform a direct word match with these keywords. A sample is retained if it meets two criteria: (1) the number of unique medical keywords $uni_k$ present exceeds a certain threshold, and (2) the proportion of the text's length occupied by the medical keywords relative to the total text length exceeds a certain threshold. Samples that do not satisfy these conditions are discarded. The mathematically equivalent expression for criteria (2) can be derived as in \eqref{kdf}:

\begin{equation} \label{kdf}
    R = \frac{\sum_k len(k) * cnt(k, T)}{len(T)}
\end{equation}

Where $k$ is the keyword in the reference list, $cnt(k, T)$ is the number of the keyword $k$ presented in the text $T$. A sample is kept if $R > thres1$ and $uni_k > thres2$.

\noindent \textbf{Answer Expanding. }Some of the data samples we collected contain answers that lack detail or are not presented in complete sentences (see example on the left of Figure \ref{preprocess}). We hypothesize that this lack of detail may hinder the interpretability and generalizability of the model. To overcome this, we employ GPT-4o-mini\footnote{\url{https://openai.com/index/gpt-4o-mini-advancing-cost-efficient-intelligence/}} to expand and elaborate on the existing answers. The prompt used for this answer expansion is provided below.

\begin{mdframed}[backgroundcolor=gray!20]
You are an experienced LANG doctor who is tasked with answering general medical questions. Here are some detailed requirements:

1. I have the answer keywords for each question. I want you to expand and write complete sentences based on the question, and answer keywords provided.

2. Your answer MUST based on the keywords, no new stuff should be present. Give me the complete answer directly in LANG.

User Input: \#\#\#Question: \{question\}, \#\#\#Answer Keywords: \{list of answer keywords\}
\end{mdframed}

\noindent \textbf{Question-Answer Pair Generation. }In some cases, corpora used for continual pretraining cannot be directly utilized for instruction fine-tuning. To address this and enhance the diversity of our dataset, we introduce a two-step processing framework to generate question-answer pairs inspired by \cite{cheng2023adapting}. First, we use GPT-4o-mini to parse each file within the corpus and produce a condensed and cleaned version of the text. Next, using the condensed text as input, we prompt GPT-4o-mini to generate two types of questions: a multiple-choice question and a short-answer question. Below, we provide the prompt used in step 1 to create the condensed text:

\begin{mdframed}[backgroundcolor=gray!20]
You’re an experienced \textbf{LANG} doctor, I have some medical texts in \textbf{LANG} that were extracted from several websites. Your task is to clean the texts and make them easy to read. Here are some detailed requirements:

1. You must ignore all links, references, and unknown characters. You MUST KEEP ALL MEDICAL-RELATED CONTENT in the texts.

2. Give the cleaned text directly without any other unnecessary words following the format of \#\#\#Cleaned Text: you cleaned text.

User Input: \#\#\#Input: \{original text\}
\end{mdframed}

Here, LANG denotes the language of the file being processed. Once the condensed text is generated, we proceed to step 2, where we use the following prompt to generate the question-answer pairs:

\begin{mdframed}[backgroundcolor=gray!20]
You’re a \textbf{LANG} medical expert tasked with creating medical questions and answers based on the short article provided. Here are the requirements:

1. For each article you read, you MUST create two types of questions: multiple-choice and short answers. One question for each type.

2. For the multiple-choice question, you MUST generate the answer choice in the following format: '\#\#\#Question: your generated question\escape{n}\escape{n}\#\#\#Options: A. optionA, B. optionB, C. optionC, D. optionD\escape{n}\escape{n}\#\#\#Rationale: your explanation\escape{n}\escape{n}\#\#\#Answer: correct answer index'

3. For the short answer question, You MUST use this format instead: '\#\#\#Question: your generated question\escape{n}\escape{n}\#\#\#Answer: your detailed answer and explanation'.

4. Multiple Choice and Short Answer questions you provided should in DIFFERENT TOPICS.

5. You MUST separate two questions by the separation symbol [SEP]. Complete the multiple-choice question first, and then switch to the short answer question.

6. You MUST strictly follow the above instructions, with NO OTHER UNNECESSARY WORDS in the output.

User Input: \#\#\#Input: \{condensed text\}
\end{mdframed}

After generating the question-answer pairs using GPT-4o-mini, we parsed them into JSON format and incorporated them into our dataset. To ensure the accuracy and quality of the generated pairs, we employed two additional models—GPT-4 \citep{achiam2023gpt} and Claude Sonnet\footnote{\url{https://www.anthropic.com/news/claude-3-5-sonnet}}—as judges. These models assessed the quality of the pairs based on three criteria: logical consistency, factual accuracy, and sound reasoning. The prompt used to evaluate the question-answer pairs is provided below:

\begin{mdframed}[backgroundcolor=gray!20]

You are an experienced and knowledgable \textbf{LANG} medical school professor. You are given a short text paired with a question-answer pair that is created based on the text. Your task is to verify the correctness of the question-answer pair. Here are some detailed requirements.

1. Based on the text provided, check whether the question-answer pair is logically consistent, factually accurate, and sound reasoning. Assign a CONTINOUS score BETWEEN 0 AND 1 for each of the criteria.

2. Only return a JSON dictionary containing THREE KEY-VALUE pairs. Here is an example that you can refer to: \{'logically consistent':0.9, 'factually accurate':0.85, 'sound reasoning':1.0\}

User Input: \#\#\#Context: \{condensed text\}\#\#\#Input: \{generated question-answer pair from GPT-4o-mni\}
\end{mdframed}

\noindent \textbf{Cross-Lingual Translation. }Translation between different languages can enrich the diversity of the dataset and can stimulate the model's full multilingual capability and interpretability \citep{lin2024crossin}. However, due to the inherent linguistic differences between languages and the complexity of medical-related samples, translating from one language to another may result in a loss or alteration of meaning. To mitigate this issue, we introduce a novel evaluation criterion called the cycle-consistency translation score to guide the translation process, inspired by the CycleGAN model \citep{zhu2017unpaired} in the computer vision domain. Formally, let $x$ be the sample in the source language $s$ and $t$ be the target language. Let $g_{s->t}$ be the translation system that translates a sample from $s$ to $t$, $g_{t->s}$ represents the reverse translation (in both cases, we use GPT-4o-mini as $g$). The cycle-consistency is defined as follows: if we translate $x$ to target language, $\Tilde{x} = g_{s->t}(x)$ and then translate $\Tilde{x}$ back to the source language, $\hat{x} = g_{t->s}(\Tilde{x})$, $x$ and $\hat{x}$ should be as similar as possible. To quantitatively evaluate the similarity, we use a combination of BLEU score \citep{papineni2002bleu} and BERT score \citep{zhang2019bertscore} as shown in \eqref{ccts}.

\begin{equation}\label{ccts}
    CCTS = \lambda_1*(\frac{1}{4}\sum_{i=1}^{4}{BLEU_{i}(x, \hat{x})}) + \lambda_2*BERT(x, \hat{x}) 
\end{equation}

Here, $BLEU_{i}$ represents BLEU-\{1,2,3,4\} score, with each component weighted equally (i.e., 0.25) and $\lambda_1, \lambda_2$ are weighting factors. Given a sample $x$ in the source language $s$, and a target language $t$, if the calculated CCTS exceeds a certain threshold, we conclude that sample $x$ can be translated into language $t$ without significant information loss, ensuring high-quality content in both languages.

We have now described all the preprocessing methods we utilized in this work. Next, we provide detailed descriptions of the multilingual dataset MMed-IFT below:

\begin{itemize}
    \item \textbf{English: }The English data in our dataset is mainly based on the combination of high-quality PMC-Llama instruction \citep{wu2024pmc}, MedInstruct \citep{zhang2023alpacare} datasets and the MMedC dataset. We randomly sample 100k data samples covering medical reading comprehension, multiple choice answering with rationale, and short answers from the PMC-Llama dataset. For MedInstruct, we performed a \textit{knowledge density filtering} based on the ratio of medical keywords represented, resulting in 34k samples left in the final version. We further performed question-answer pair generation on 3k randomly selected samples from MMedC. The total number of samples for English data is around 135k.
    
    \item \textbf{Chinese: }The Chinese data in our dataset is derived from a combination of the Huatuo Encyclopedia, Huatuo Knowledge Graph QA dataset \citep{li2023huatuo}, and the MMedC dataset \citep{qiu2024towards}. For the Knowledge Graph QA dataset, we applied answer expansion to enhance the detail and completeness of the answers for each question. Regarding the MMedC dataset, which was originally designed for continual pretraining, we conducted question-answer pair generation to convert it into an instruction fine-tuning format on 30k randomly sampled text files. The total number of samples for the Chinese data is around 80k. 
    
    \item \textbf{Korean: }We hand-crafted the Korean data in our dataset. We first randomly sampled 10k data from both the English and the Chinese datasets mentioned above, and then we performed English-Korean and Chinese-Korean translations. We retained only the samples with a cycle-consistency translation score (CCTS) greater than 0.8. Additionally, we processed several Korean books and extracted 3.2k question-answer pairs from them. The total number of samples for the Korean data is around 20k. 
    
    \item \textbf{Japanese: }The Japanese data in our dataset is composed of hand-crafted samples and the MMedC dataset \citep{qiu2024towards}. The hand-crafting process was similar to that used for the Korean dataset, except we extracted approximately 1k question-answer pairs from a Japanese book. For MMedC, we generated question-answer pairs to convert them into an instruction fine-tuning format on 10k randomly selected samples. The total number of samples for the Japanese data is around 20k. 
    
    \item \textbf{French: }The French data in our dataset is sourced from the Apollo Corpus \citep{wang2024apollo} and the MMedC dataset \citep{qiu2024towards}. We directly utilized the samples from the Apollo Corpus. For the MMedC dataset, we generated question-answer pairs to convert them into an instruction fine-tuning format on 5k randomly selected samples. The total number of samples for the French data is around 20k. 
    
    \item \textbf{Spanish: }We applied the same procedures to the Apollo Corpus \citep{wang2024apollo} and the MMedC \citep{qiu2024towards} dataset as described for the French data to construct our Spanish dataset. The total number of samples for the Spanish data is around 20k. 
    
\end{itemize}

We further collected a more task-specific dataset, \textbf{MMed-IFT-MC}, containing \textit{exclusively} Medical License Examination (MLE)-style multiple-choice questions across six languages of around 50k samples. This dataset was derived from MMedBench \citep{qiu2024towards} and KorMedMCQA \citep{kweon2024kormedmcqa} with rationale, as current medical benchmarks predominantly consist of this type of multiple-choice questions. We adopted the same train-test split as specified in the original works. To avoid data leakage, we randomly cherry-picked 100 samples for each language from the MMed-IFT dataset and verified that these samples were not included in both the MMed-IFT-MC and test set.

\subsection{Two-Stage IFT Approach}

Training large language models (LLMs) in the medical domain typically involves continual pre-training on a medical corpus, such as medical books, papers, and related documents. However, recent studies \citep{wang2024apollo,chen2023huatuogpt,cheng2023adapting} have demonstrated that while pre-training on such corpora provides the model with general domain knowledge, it significantly impairs its ability to handle prompt-based question-answering tasks. Furthermore, continual pre-training demands substantial computational resources \citep{qiu2024towards,chen2023meditron,wu2024pmc}, which may not be feasible in certain situations. To overcome these limitations, we propose a two-stage, parameter-efficient instruction fine-tuning approach. We hypothesize that by fine-tuning the LLM on our carefully curated MMed-IFT dataset, which consists of diverse question-answer pairs spanning various medical topics, the model will acquire sufficient general medical knowledge. Hence, in the first stage, we inject this medical knowledge into the LLM using the MMed-IFT dataset through LoRA fine-tuning \citep{hu2021lora}. In the second stage, we perform task-specific fine-tuning on MLE-style multiple-choice questions using the MMed-IFT-MC dataset, again applying LoRA fine-tuning to improve the model’s multilingual interpretability and maximize its performance on existing medical benchmarks. Note that after the first stage, we merge the LoRA weights with the base LLM to ensure it retains comprehensive general medical knowledge. This merged LLM is then used for the second stage, further refining its capabilities for specific medical tasks.

\section{Experiments}

All programs were implemented using PyTorch, HuggingFace Transformers \citep{wolf2019huggingface}, and HuggingFace PEFT \citep{peft}. We conducted all experiments on a single NVIDIA GeForce RTX 4090 with 25GB of memory, using Llama3-8B \citep{dubey2024llama} as the base model.

In the first stage, we applied a quantized version of DoRA \citep{liu2024dora} with rank = 32, alpha = 16, and dropout = 0.05, targeting all linear layers for parameter-efficient fine-tuning (PEFT). The model was fine-tuned for 2 epochs with a batch size of 1, an initial learning rate of 5e-5 (decayed to 0 using a cosine scheduler), a warmup ratio of 0.2, and gradient accumulation every 4 steps.

In the second stage, we used QLoRA \citep{dettmers2024qlora} for fine-tuning, with the same hyperparameters as the first stage except for rank = 16, alpha = 8, and an initial learning rate of 2e-5.

\noindent \textbf{Evaluations: }All evaluations are done using the \texttt{LM-Evaluation-Harness}\footnote{https://github.com/EleutherAI/lm-evaluation-harness} framework. We evaluate our model on various medical benchmarks, including the USMLE Self-assessment Step 1, 2, 3 datasets from ClinicalCamel \citep{toma2023clinical}, along with the official test splits of MedQA-4-Options \citep{jin2021disease} and MedMCQA \citep{pal2022medmcqa} for English benchmarks. For the Korean benchmark, we utilize KorMedMCQA \citep{kweon2024kormedmcqa}. For Chinese, Spanish, French, and Japanese benchmarks, we use the same test sets as described in \cite{qiu2024towards}. For KorMedMCQA \citep{kweon2024kormedmcqa} dataset, we use the exact match criterion as the evaluation metric following the original paper while accuracy is used for all other test datasets. For comparison, we select various models covering three main types: Open-sourced LLMs (base models), models that have undergone continual pre-training and instruction fine-tuning on a medical corpus, and models that are only instruction fine-tuned on the medical corpus. All selected models are of similar sizes (7B/8B/13B) to our base model for a fair comparison. Below, we introduce the baseline LLMs used in our work:

\begin{itemize}
    \item Llama2-7B \citep{touvron2023llama} and Llama3-8B \cite{dubey2024llama}: The Llama series is a collection of open-source large language models (LLMs) developed by Meta. Llama2 represents the previous generation, while Llama3 is one of the recent releases. These models are widely recognized as the most powerful open-source LLMs for English. Although Llama2’s substantial pre-training data likely includes samples from other languages, allowing it to perform reasonably well in multilingual scenarios, Llama3 is explicitly trained on multilingual datasets. This makes Llama3 highly effective in multilingual tasks, which is why we chose it as our base model.
    \item PMC-Llama-13B \citep{wu2024pmc}: PMC-Llama is an English-centric medical LLM that undergoes continual pre-training and instruction fine-tuning on a medical corpus using Llama as the base model.
    
    \item Meditron-7B \citep{chen2023meditron}: Meditron is an open-source biomedical LLM focused on English, continually pre-trained on Llama2 using approximately 45 billion English tokens.
    
    \item Alpacare-7B \citep{zhang2023alpacare}: Alpacare is a monolingual, open-source LLM that is instruction fine-tuned on a 54k hand-crafted English medical dataset.
    
    \item MedAlpaca-7B \citep{han2023medalpaca}: MedAlpaca is a specialized, open-source English medical LLM, instruction fine-tuned on Llama using over 160k English medical records.
    
    \item Med42-8B \citep{christophe2024med42}: Med42 is a set of open-source clinical LLMs based on Llama3 and Llama3.1, instruction fine-tuned and performance-aligned using direct policy optimization \citep{rafailov2024direct}. We used the 8B version of Med42.
    
    \item MMed-Llama3-8B \citep{qiu2024towards}: MMed-Llama3 is a state-of-the-art \textbf{multilingual} medical LLM covering English, Chinese, Japanese, French, Spanish, and Russian. It is continually pre-trained on 25.5 billion tokens across all six languages, followed by instruction fine-tuning on curated multiple-choice questions. An English-specific variant, MMed-Llama3-EnIns, is released for fair comparison on English benchmarks. We use both the MMed-Llama3 and MMed-Llama3-EnIns as the baselines.
    
    \item Apollo-7B \citep{wang2024apollo}: Apollo is a state-of-the-art multilingual medical LLM covering English, Chinese, French, Spanish, Arabic, and Hindi. It undergoes a mixed training process on the pre-training and instruction fine-tuning datasets. The pre-training dataset has been reformatted to the question-answer pairs. We used the 7B variant as the baseline.
    
\end{itemize}

\section{Results}

We provide a preliminary analysis of the results, as our work is still ongoing. Table \ref{main_res} presents the main results for the English benchmarks. Our results for \textbf{TS-Llama3-MMed-IFT-En} are based on using only the English portion of the MMed-IFT dataset for the first stage and the entire MMed-IFT-MC dataset for the second stage in a pilot experiment. Training on the whole MMed-IFT dataset is time-consuming, and we do not yet have those results. Despite this limitation, our method demonstrates significant improvements over all other instruction fine-tuned models and achieves performance comparable to the state-of-the-art MMed-Llama3-8B-EnIns (underscored in Table \ref{main_res}).

Compared to our Llama3-MMed-IFT-MC model, which was only instruction fine-tuned on the MMed-IFT-MC dataset, our TS-Llama3-MMed-IFT-En consistently outperforms most English benchmarks, further validating the effectiveness of general medical knowledge injection during the first stage of IFT. This highlights the strength of our two-stage IFT framework: the first stage successfully injects clinical knowledge into the model, while the second stage further enhances the model's reasoning abilities on specific tasks.

\begin{mdframed}
Key Takeaways:
\begin{itemize}
    \item The general medical knowledge injection in the first stage is necessary to improve the performance of LLM.
    \item Our preprocessed English dataset is of high quality and the question-answer generation is effective.
\end{itemize}
\end{mdframed}

\begin{table}[ht]
\centering
\resizebox{\textwidth}{!}{\begin{tabular}{|cccccc|}
\hline
\multicolumn{1}{|c|}{Method}                                                                              & \multicolumn{1}{c|}{USMLE Step 1}  & \multicolumn{1}{c|}{USMLE Step 2} & \multicolumn{1}{c|}{USMLE Step 3} & \multicolumn{1}{c|}{MedQA-4-Options} & MedMCQA \\ \hline
\multicolumn{6}{|c|}{\colorbox{lightgray}{Open-sourced Base Model}}  \\ \hline

\multicolumn{1}{|c|}{Llama-2-7B}                                                                          & \multicolumn{1}{c|}{22.3}              & \multicolumn{1}{c|}{22.9}             & \multicolumn{1}{c|}{22.1}             & \multicolumn{1}{c|}{32.6}                &  31.3       \\ \hline
\multicolumn{1}{|c|}{Llama-3-8B}                                                                          & \multicolumn{1}{c|}{46.8}          & \multicolumn{1}{c|}{43.1}         & \multicolumn{1}{c|}{44.3}         & \multicolumn{1}{c|}{53.7}            & 49.3    \\ \hline
\multicolumn{6}{|c|}{\colorbox{lightgray}{Continual Pretraining + IFT}}                                                                                                                                                                                                                      \\ \hline
\multicolumn{1}{|c|}{PMC-Llama-13B}                                                                       & \multicolumn{1}{c|}{34.0}          & \multicolumn{1}{c|}{30.3}         & \multicolumn{1}{c|}{41.8}         & \multicolumn{1}{c|}{50.3}            & 49.3    \\ \hline
\multicolumn{1}{|c|}{Meditron-7B}                                                                         & \multicolumn{1}{c|}{24.5}              & \multicolumn{1}{c|}{17.4}             & \multicolumn{1}{c|}{20.5}             & \multicolumn{1}{c|}{29.7}                &  29.3       \\ \hline
\multicolumn{1}{|c|}{MMed-Llama3-8B}                                                                      & \multicolumn{1}{c|}{52.1} & \multicolumn{1}{c|}{43.1}         & \multicolumn{1}{c|}{50.8}         & \multicolumn{1}{c|}{55.3}            & 52.2    \\ \hline
\multicolumn{1}{|c|}{MMed-Llama3-8B-EnIns}                                                                & \multicolumn{1}{c|}{\underline{57.4}}     & \multicolumn{1}{c|}{\underline{60.5}}             & \multicolumn{1}{c|}{\underline{68.8}}             & \multicolumn{1}{c|}{\underline{61.4}}                &  \underline{59.1}        \\ \hline
\multicolumn{6}{|c|}{\colorbox{lightgray}{IFT Only}}                                                                                                                                                                                                                                     \\ \hline
\multicolumn{1}{|c|}{Llama2-Alpacare-7B}                                                                  & \multicolumn{1}{c|}{19.1}          & \multicolumn{1}{c|}{21.1}         & \multicolumn{1}{c|}{18.8}         & \multicolumn{1}{c|}{29.9}            & 34.3    \\ \hline
\multicolumn{1}{|c|}{MedAlpaca-7B}                                                                        & \multicolumn{1}{c|}{30.8}          & \multicolumn{1}{c|}{25.7}         & \multicolumn{1}{c|}{34.4}         & \multicolumn{1}{c|}{41.8}            & 36.3    \\ \hline
\multicolumn{1}{|c|}{Llama3-Med42-8B}                                                                         & \multicolumn{1}{c|}{54.3}              & \multicolumn{1}{c|}{54.1}             & \multicolumn{1}{c|}{59.8}             & \multicolumn{1}{c|}{56.5}                &   54.0      \\ \hline
\multicolumn{1}{|c|}{Apollo-7B}                                                                           & \multicolumn{1}{c|}{34.0}              & \multicolumn{1}{c|}{41.3}             & \multicolumn{1}{c|}{41.8}             & \multicolumn{1}{c|}{47.4}                &  44.1       \\ \hline
\multicolumn{1}{|c|}{Llama3-Alpacare-7B}                                                                  & \multicolumn{1}{c|}{47.9}          & \multicolumn{1}{c|}{44.0}         & \multicolumn{1}{c|}{45.1}         & \multicolumn{1}{c|}{54.5}            & 49.2   \\ \hline
\multicolumn{1}{|c|}{\begin{tabular}[c]{@{}c@{}}Llama3-MMed-IFT-MC\\ (Ours)\end{tabular}}                 & \multicolumn{1}{c|}{55.3}          & \multicolumn{1}{c|}{\textbf{52.3}}         & \multicolumn{1}{c|}{57.4}         & \multicolumn{1}{c|}{59.7}            & 53.6    \\ \hline
\multicolumn{1}{|c|}{\begin{tabular}[c]{@{}c@{}}TS-Llama3-MMed-IFT-En\\ (Ours)\end{tabular}}             & \multicolumn{1}{c|}{\textbf{56.4}}              & \multicolumn{1}{c|}{\textbf{52.3}}             & \multicolumn{1}{c|}{\textbf{67.2}}             & \multicolumn{1}{c|}{\textbf{62.4}}                &    \textbf{57.0}     \\ \hline
\end{tabular}}
\caption{Overall comparison of both the open and closed source LLMs on English medical benchmarks. We separate the table into three sections: base models; models that are further continually pre-trained and instruction fine-tuned on the medical corpus; and models that are only instruction fine-tuned on the medical corpus. IFT is short for Instruction Fine-tuned. Bold numbers represent the best performance.}
\label{main_res}
\end{table}

Next, we provide some preliminary insights into the multilingual results in Table \ref{multilin_res}. We selected the Llama3 base model and two variants of our models: one fine-tuned solely on the MMed-IFT-MC dataset, and the other fine-tuned on the English portion of the MMed-IFT dataset in the first stage, followed by further fine-tuning with the MMed-IFT-MC dataset in the second stage. Our observations indicate that injecting general medical knowledge in the first stage is crucial for improving performance in multilingual settings.

\begin{mdframed}
Key Insights:
\begin{itemize}
    \item Injecting general medical knowledge in the first stage proves effective in multilingual settings.
    \item Injecting general medical knowledge in English can also enhance performance across other languages.
\end{itemize}
\end{mdframed}

\noindent \textbf{Note:} We will work on the first-stage fine-tuning using the entire MMed-IFT dataset and update the results in the future.

\begin{table}[ht]
\centering
\resizebox{\textwidth}{!}{\begin{tabular}{|c|c|c|c|c|c|}
\hline
Method                                                                  & CMLE                  & KrMedMCQA             & JMLE                  & FrMedMCQA             & HeadQA-ES             \\ \hline
Llama-3-8B                                                              & 38.1                  & 38.4                  &   25.6                    & 29.4                  & 44.7                  \\ \hline
\begin{tabular}[c]{@{}c@{}}Llama3-MMed-IFT-MC\\ (Ours)\end{tabular}     & 46.5                  & 44.8                  &    30.7                   & 46.8                  & 46.5                  \\ \hline
\begin{tabular}[c]{@{}c@{}}TS-Llama3-MMed-IFT-En\\ (Ours)\end{tabular} &  \textbf{59.3}                     &      \textbf{46.0}                 &    \textbf{43.2}                   &     \textbf{55.6}                  &       \textbf{61.9}                \\ \hline
\end{tabular}}
\caption{Overall comparison of the selected LLMs on Multilingual medical benchmarks. CMLE is short for Chinese MLE and JMLE is short for Japanese MLE. Bold numbers represent the best performance.}
\label{multilin_res}
\end{table}

\section{Conclusion}

In this work, we address the limitations of relying on a single instruction fine-tuning dataset for adapting large language models (LLMs) to the medical domain, particularly in multilingual settings. Continual pretraining, while effective, often requires substantial computational resources, making it impractical in many cases. To overcome these challenges, we propose a two-stage instruction fine-tuning framework, leveraging our newly constructed MMed-IFT and MMed-IFT-MC datasets. We plan to refine and update our multilingual performance results and conduct ablation studies to further validate our method in the future.


\bibliography{iclr2025_conference}
\bibliographystyle{iclr2025_conference}


\end{document}